\documentclass{article}
\usepackage{iclr2026_conference,times}

\usepackage{amsmath,amsfonts,bm}

\def\eqref#1{equation~\ref{#1}}

\def\1{\bm{1}}

\DeclareMathAlphabet{\mathsfit}{\encodingdefault}{\sfdefault}{m}{sl}
\SetMathAlphabet{\mathsfit}{bold}{\encodingdefault}{\sfdefault}{bx}{n}

\usepackage{amsthm}
\newtheorem{theorem}{Theorem}[section]   %

\newtheorem{proposition}[theorem]{Proposition}

\theoremstyle{definition}

\usepackage{amsmath,amssymb}
\usepackage{booktabs}
\usepackage{multirow}
\usepackage{array}
\usepackage{wrapfig}
\usepackage{graphicx}
\usepackage{xcolor}
\definecolor{costred}{RGB}{190,45,45}
\usepackage{tikz}
\usetikzlibrary{arrows.meta,positioning,calc,fit,backgrounds}
\usepackage{microtype}
\usepackage{enumitem}
\usepackage{float}
\usepackage{xspace}
\usepackage{hyperref}
\usepackage{url}

\definecolor{AdvisorBlue}{HTML}{315B8A}
\definecolor{AdvisorTeal}{HTML}{2A8C82}
\definecolor{AdvisorGold}{HTML}{D49A3A}
\definecolor{AdvisorCoral}{HTML}{C95D63}
\definecolor{AdvisorInk}{HTML}{25313C}
\definecolor{AdvisorLight}{HTML}{F3F6F8}

\title{Don’t Solve, Just Compare: Tiny Advisors\\ for Runtime Intervention in LLM Agents}

\author{Yanze Jiang\thanks{Equal contribution.}\;\,, 
Mingxuan Li\footnotemark[1]\;\,, 
Yuhao Wang, 
Shengfang Zhai\thanks{Corresponding authors.}\;\,,
Jiaheng Zhang\footnotemark[2] 
\\  
National University of Singapore 
\\  
\texttt{\{yanzejiang, e1553202, wangyuhao\}@u.nus.edu} \\ 
\texttt{\{shengfang.zhai, jhzhang\}@nus.edu.sg}}

\newcommand{\name}{\textsc{COTA}\xspace}

\iclrfinalcopy
\begin{document}

\maketitle
\lhead{}
\begin{abstract}

LLM agents are emerging as an important paradigm for real-world tasks that require reasoning, tool use, and sequential decision-making. As these agents operate over longer horizons, runtime intervention offers a way to improve reliability without retraining the underlying actor. Effective intervention must provide a useful direction for recovery besides a warning. Existing approaches often rely on an expert solver or a critic that generates task-specific corrections, incurring either the cost of another capable solver or the capacity demands of a task-capable critic. 
We introduce \textbf{C}omparison-\textbf{O}nly \textbf{T}iny \textbf{A}dvisor (\textbf{\name}) for constructive runtime intervention, which reduces the learned intervention role to local action comparison. A lightweight comparator judges the actor's proposal against available alternatives, and preferred alternatives are returned as non-binding advice for replanning. The comparator is trained from same-prefix counterfactual branches.  
Across WebShop, ALFWorld, and $\tau^3$-Retail with three LLM actors, \name instantiated with a 0.5B comparator consistently improves the original actor and achieves the strongest overall performance--cost trade-off among the compared methods. 
These results suggest that effective runtime intervention need not itself be a task-solving problem: the intervention role can be separated from task solving and handled by a lightweight model specialized for local comparison.
\end{abstract}

\section{Introduction}
\label{sec:introduction}

Large language model (LLM) agents are increasingly used for long-horizon tasks that require sequential reasoning, tool use, and interaction with external environments~\citep{yao2022react,shinn2023reflexion}. In such settings, decisions are consequential: a locally poor action can alter the environment, consume a limited interaction budget, and make subsequent recovery substantially harder. \emph{Runtime intervention} provides a practical way to improve reliability without retraining or replacing the underlying actor, by monitoring an ongoing trajectory and redirecting execution when needed~\citep{vasudev2026accurate,zhang2026calibration,jiang2026asymmetric}. Yet effective runtime intervention is not merely a failure-detection problem. Recent work shows that even accurate failure predictors can reduce end-to-end performance by disrupting trajectories that would otherwise succeed~\citep{vasudev2026accurate}, and argues that effective runtime control should depend on whether an available intervention improves the downstream outcome, rather than on continuation risk alone~\citep{zhang2026calibration}. 
Effective runtime intervention therefore requires more than detecting problematic decisions. It must also provide a recovery signal that improves what the agent does next.

Existing approaches pursue this goal through several routes. Self-reflection asks the actor to reconsider its own proposal, but without an external signal it can preserve or even amplify the same reasoning error~\citep{shinn2023reflexion,huang2024selfcorrect,kamoi2024selfcorrection}. Expert-handoff methods intervene by transferring control to a stronger solver when the current trajectory appears problematic~\citep{zhang2026calibration,ong2025routellm,shen2024collm}; while potentially effective, they require invoking another capable model. Critic-based methods instead keep the actor in control and use a separate model to generate task-specific corrective feedback~\citep{jiang2026asymmetric}, but producing useful corrections still demands substantial task understanding from the critic, making it difficult to keep lightweight~\citep{chen2024frugalgpt,ong2025routellm}. 
More broadly, reward-model-based inference-time search can improve agent performance by generating multiple complete trajectories and selecting the highest-scored one~\citep{xia2025agentrm,choudhury2025process,xi2026agentprm}. Its effectiveness depends on reliable ranking, while larger search budgets require more trajectory rollouts and therefore higher inference cost.
This motivates the question we study in this work: Can effective constructive runtime intervention be realized by a separate, lightweight intervention mechanism while requiring substantially weaker learned capabilities than task solving or correction generation?

Doing so presents two challenges. First, constructive intervention must provide more than a binary warning: when the actor proposes a poor action, it should offer a useful direction for replanning~\citep{bai2022constitutional,gou2023critic}. Yet asking the tiny learned intervention module itself to discover that direction would reintroduce open-ended task solving~\citep{lin2024criticbench,chen2024selfdebug}. Second, intervention should reflect the long-term consequence of the current decision rather than its surface plausibility~\citep{sutton1988temporal,pignatelli2024credit,xiong2024watch}. Learning accurate absolute values for arbitrary state--action pairs is particularly demanding in long-horizon agent tasks~\citep{arjona2019rudder,gehring2022dense,parisi2022visitation,xi2026agentprm}, while comparing actions observed under different trajectories mixes the effect of the action with differences in history and continuation policy~\citep{oberst2019counterfactual,jiang2016doubly,thomas2015highconfidence}.

These challenges motivate a weaker primitive: comparison. An action need not be certified as globally good to determine that the actor's current proposal is locally weak. If the proposed action is repeatedly outperformed by plausible executable alternative actions, then the proposed action is likely a poor local choice, even when none of those alternative actions is itself optimal. Based on this observation, we propose \textbf{C}omparison-\textbf{O}nly \textbf{T}iny \textbf{A}dvisor (\textbf{\name}), a comparison-only framework for constructive runtime intervention. 
First, to make comparison constructive, 
at each decision point, we obtain a small set of alternative actions from a reference candidate mechanism whose construction is left configurable, and a tiny comparator performs only one primitive: predicting which of two actions from the same state leads to the better continuation. 
Repeated pairwise judgments estimate how frequently the actor's proposal is dominated by its alternatives. When this candidate-relative domination is sufficiently strong, the winning alternatives are returned as non-binding advice and the actor replans before execution.
Second, to learn this long-horizon comparison without absolute value estimation or cross-trajectory confounding, 
we use same-prefix counterfactual branching, inspired by~\cite{zhang2026calibration}. Starting from an identical environment state, we vary only the branch-point action and then return control to the same frozen actor. 
The resulting sibling trajectories provide pairwise supervision that directly matches the comparison required at runtime.
The comparator therefore neither generates corrective actions nor estimates absolute action values. It only compares.

We evaluate \name on WebShop~\citep{yao2022webshop}, ALFWorld~\citep{shridhar2020alfworld}, and $\tau^3$-Retail~\citep{yao2024tau}, using Qwen3-8B, Qwen3.6-35B-A3B, and DeepSeek-V4-Flash as LLM agent actors.
Across all evaluation combinations, despite using only a 0.5B comparator, \name consistently improves the original actor while maintaining a favorable performance--cost trade-off.
The gains extend to substantially stronger actors, showing that effective constructive intervention can be separated from task solving and realized by a lightweight model specialized for local comparison, while leaving final replanning to the actor. 
Our contributions are threefold:
\begin{itemize}[leftmargin=0.9em]
\item We formulate \emph{constructive runtime intervention} and show that the capability required for its intervention role can be reduced from open-ended task solving or correction generation by a strong LLM to candidate-relative comparison, which can be handled by a separate, lightweight model. 

\item We propose \name, a comparison-only framework that aggregates pairwise judgments through a statistically interpretable Monte Carlo gate. We further construct reliable pairwise supervision from same-prefix counterfactual trajectories, directly matching the comparison required at runtime.

\item We evaluate \name across three interactive environments and three actor families. A 0.5B comparator improves all nine actor--environment combinations, transfers without retraining to substantially stronger actors, and achieves a favorable performance--cost trade-off.

\end{itemize}

\section{Problem Formulation}
\label{sec:problem}

\noindent \textbf{Agent trajectory.}
Following the standard formulation of interactive LLM agents~\citep{yao2022react} and sequential decision making under partial observability~\citep{kaelbling1998planning}, we consider a frozen LLM agent $\pi$ interacting with an environment to accomplish a task $x$. 
We refer to the frozen LLM policy $\pi$ as the \textit{actor}. 
At each step $t$, it observes $o_t$, produces a reasoning trace or plan $p_t$, and proposes an environment action $a_t$. 
If $a_t$ is executed, the environment returns the next observation $o_{t+1}$. We denote the trajectory prefix before executing $a_t$ as:
\begin{equation}
\tau_t
=
(o_0,p_0,a_0,\ldots,o_{t-1},p_{t-1},a_{t-1},o_t,p_t),
\label{eq:trajectory}
\end{equation}
and let $s_t=(x,\tau_t)$ denote the actor-visible decision state. Thus, the proposed action is sampled as $a_t\sim\pi(\cdot\mid s_t)$. Here $s_t$ denotes the information available to the agent at decision time, rather than necessarily full environment state.

\noindent \textbf{Runtime intervention.}
In this paper, we consider runtime intervention~\citep{wang2026agentspec,chen2025shieldagent,wang2026probguard} at the \emph{pre-execution} stage. The agent has already proposed $a_t$, but the action has not yet been sent to the environment. 
We follow the general human-intervention and shielding paradigm that an external mechanism monitors a proposed action and may prevent or modify it before execution~\citep{saunders2018trial,alshiekh2018shielding}. 
Let $g_t\in\{0,1\}$ denote the intervention decision. 
The original proposal is executed if $g_t=0$ and withholds if $g_t=1$. 
Our definition does not specify what happens after an action is withheld.

For any action $a$ available at $s_t$, we define its action value~\citep{watkins1992qlearning,sutton2018reinforcement} under the actor $\pi$ as: 
\begin{equation}
Q^\pi(s_t,a)
=
\mathbb{E}
\left[
R(\tau)
\mid
s_t,; a_t=a,; \pi\ \text{thereafter}
\right],
\label{eq:qpi}
\end{equation}
where $R(\tau)$ is the return of the completed trajectory. $Q^\pi(s_t,a)$ is the expected return from taking $a$ at $s_t$ and following $\pi$ thereafter.

\noindent \textbf{Constructive runtime intervention.}
Compared with prior runtime monitoring and intervention methods \citep{xiang2025guardagent,wang2026probguard}, which primarily determine whether a proposed action $a_t$ should be allowed to proceed or interrupted, without necessarily providing a concrete direction for recovery, 
we emphasize that a more constructive runtime intervention should provide an advisory action $a_t^{\mathrm{adv}}$ to the actor $\pi$ when withholding $a_t$, thereby guiding subsequent action selection. 
The advice is non-binding: the actor replans after receiving it and produces a new proposal $a_t'$. 
We call the intervention constructive when the advice offers a better direction: %
\[Q^\pi(s_t,a_t^{\mathrm{adv}})
>
Q^\pi(s_t,a_t).
\]
The desired end-to-end outcome is an improved proposal: %
\[
Q^\pi(s_t,a_t')
>
Q^\pi(s_t,a_t).
\]
We call the separate role that performs this advice-producing intervention an \emph{advisor}.
Our goal is to enable such intervention with a lightweight advisor that need not solve the task independently. 
\section{Methodology}
\label{sec:method}
In this section, we present \name in four parts. We first introduce the comparison intuition underlying our approach in Section~\ref{sec:method_intuition}, and then describe the \name runtime intervention pipeline in Section~\ref{sec:tiny_advisor_pipeline}. We next describe the design of comparator training using same-prefix branches in Section~\ref{sec:comparator_training}, and finally analyze the statistical accuracy of \name's intervention gate in Section~\ref{sec:gate_accuracy}.

\subsection{Intuition}
\label{sec:method_intuition}

Directly estimating whether $\pi$'s proposal action $a_t$ is ``good'' through \(Q^\pi(s_t,a_t)\) is unnecessarily difficult for constructive runtime intervention. It is sufficient to determine whether \(a_t\) is poor compared with plausible alternatives available at the same decision point. Let \(\mu(\cdot\mid s_t)\) denote the distribution over executable actions induced by a reference candidate mechanism. We define the \emph{candidate-relative domination rate}: 
\begin{equation}
\rho_\mu(s_t,a_t)
=
\Pr_{A\sim\mu(\cdot\mid s_t)}
\left(
Q^\pi(s_t,A)
>
Q^\pi(s_t,a_t)
\right).
\label{eq:rho}
\end{equation}
This quantity has a simple rank interpretation. If
\(\rho_\mu(s_t,a_t)=0.1\), only \(10\%\) of candidates sampled from \(\mu\) outperform the proposal; if
\(\rho_\mu(s_t,a_t)=0.8\), the proposal $a_t$ is dominated by \(80\%\) of the candidate distribution $\mu$. Equivalently, \(1-\rho_\mu\) gives $a_t$'s quantile in $\mu$, up to ties. Importantly, \(\mu\) need not be an expert policy or contain the globally optimal action. If \(\rho_\mu\) is small, $a_t$ already compares favorably with ordinary alternatives; conversely, a large \(\rho_\mu\) provides evidence that even ordinary alternatives frequently improve upon it. Moreover, the alternatives that defeat $a_t$ naturally serve as advice for replanning. 
This motivates a much narrower learned primitive for constructive runtime intervention: predicting which of two actions from the same state leads to the better continuation.

\subsection{Comparison-Only Tiny Advisor}
\label{sec:tiny_advisor_pipeline}

Based on this statistical insight, we propose \name, 
a runtime intervention framework that realizes the advisor using a single learned primitive: pairwise action comparison.
At step \(t\), $\pi$ first proposes \(a_t\sim\pi(\cdot\mid s_t)\). We then obtain \(K\) executable alternative actions from $\mu$, i.e., \(\mathcal{A}_t
=
\{A_1,\ldots,A_K\}\), \(
A_i\sim\mu(\cdot\mid s_t).
\)
A tiny comparator \(C_\theta\) judges each candidate $A_i$ only relative to $a_t$. \(C_\theta\)'s target is: 
\begin{equation}
C_\theta(s_t,a_A,a_B)
\approx
\mathbb{I}
\left[
Q^\pi(s_t,a_A)
>
Q^\pi(s_t,a_B)
\right].
\label{eq:comparator_target}
\end{equation}
For each \(A_i\), we define: 
\begin{equation}
\widetilde B_i
=
C_\theta(s_t,A_i,a_t)
\in \{0,1\},
\label{eq:pairwise_decision}
\end{equation}
where \(\widetilde B_i=1\) indicates that \(A_i\) is predicted to outperform $a_t$. 
\name rejects \(a_t\) when at least \(R\) of the \(K\) alternatives defeat it: 
\begin{equation}
\sum_{i=1}^{K}\widetilde B_i
\geq R.
\label{eq:runtime_reject_rule}
\end{equation}
Otherwise, \(a_t\) is executed unchanged. When the gate fires, the predicted winners are ranked using \(C_\theta\), and the highest-ranked candidate \(a_t^{\mathrm{adv}}\) is returned to $\pi$ as non-binding advice. $\pi$ then replans from the same $s_t$ conditioned on this advice and produces a new proposal \(a_t'\), which is reviewed before execution. 
The comparator \(C_\theta\) is therefore the only learned primitive required by \name: it neither generates corrective actions nor estimates absolute action values, but only performs the local comparison in Eq.~\ref{eq:comparator_target}. 
$\pi$ remains responsible for planning. 
\name only assumes access to such a $\mu$ and is agnostic to how \(\mu\) is constructed. We describe the concrete $\mu$ mechanisms used in our experiments in Section~\ref{sec:implementation}. 
We detail the statistical accuracy of this candidate-relative gate in Sec.~\ref{sec:gate_accuracy}. 

\subsection{Comparator Training}
\label{sec:comparator_training}

Training \(C_\theta\) requires supervision for the relative continuation quality of two actions taken from the same state. Returns from unrelated trajectories are unsuitable for this purpose because they conflate the effect of the action with differences in preceding histories and subsequent behavior. We therefore construct training data using \emph{same-prefix counterfactual branches},
following the classical idea of Monte-Carlo rollout evaluation from a
shared decision state~\citep{tesauro1996online,bertsekas1999rollout}.

At a sampled decision state \(s_t\), we restore the same environment state, execute a branch-point action \(a\), and then return control to the same frozen actor \(\pi\). Let \( Y(s_t,a)
=
R(\tau)\), \(
\tau\sim P(\cdot\mid s_t,a,\pi)
\) 
denote the resulting branch return. By the definition of $Q^\pi(s_t,a)$ in Eq.~\ref{eq:qpi}: 
\begin{equation}
\mathbb{E}\!\left[Y(s_t,a)\right]
=
Q^\pi(s_t,a).
\label{eq:branch_expectation}
\end{equation}
Hence, \(
\widehat Q_M(s_t,a)
=
\frac{1}{M}
\sum_{m=1}^{M}
Y_m(s_t,a)
\)
is a Monte Carlo estimate of \(Q^\pi(s_t,a)\) with \(M\) independent continuations. 
For sibling actions \(a_A\) and \(a_B\), their empirical branch returns directly provide pairwise supervision:
\begin{equation}
\widehat B(s_t,a_A,a_B)
=
\mathbb{I}
\left[
\widehat Q_M(s_t,a_A)
>
\widehat Q_M(s_t,a_B)
\right].
\label{eq:pairwise_label}
\end{equation}
We fine-tune \(C_\theta\) to predict these pairwise labels from the actor-visible state and the two candidate actions. Because both branches share the same prefix and continuation actor, the comparison isolates the downstream consequence of changing the branch-point action. Details of label construction and comparator implementation are provided in Section~\ref{sec:implementation}.

\subsection{Statistical Accuracy}
\label{sec:gate_accuracy}

In this part, we analyze the statistical accuracy of \name's candidate-relative gate defined in~Sec.\ref{sec:tiny_advisor_pipeline}. 
Suppose first that the true pairwise relation were observable, and define \(
B_i
=
\mathbb{I}
\left[
Q^\pi(s_t,A_i)
>
Q^\pi(s_t,a_t)
\right].
\)
For \(K\) sampled candidates, let: 
\begin{equation}
\widehat\rho_K
=
\frac{1}{K}
\sum_{i=1}^{K} B_i,
\qquad
\mathbb{E}\!\left[\widehat\rho_K\right]
=
\rho_\mu(s_t,a_t).
\label{eq:oracle_rho}
\end{equation}
Thus, \(\widehat\rho_K\) is a Monte Carlo estimate of the candidate-relative domination rate
\(\rho_\mu(s_t,a_t)\) in Eq.~\ref{eq:rho}.
Further, requiring \(R\) winners in Eq.~\ref{eq:runtime_reject_rule} corresponds to testing whether \(\rho_\mu(s_t,a_t)\ge \frac{R}{K}\), using \(\widehat{\rho}_K\) as its Monte Carlo estimate.
That is, whether $a_t$ is dominated by at least an \(\frac{R}{K}\) fraction of $\mu$.

At deployment, we replace the oracle comparisons \(B_i\) with the learned predictions \(\widetilde B_i\): 
\begin{equation}
\widehat\rho_{\theta,K}
=
\frac{1}{K}
\sum_{i=1}^{K}
\widetilde B_i,
\qquad
g_\theta
=
\mathbb{I}
\left[
\widehat\rho_{\theta,K}
\geq
\frac{R}{K}
\right].
\label{eq:learned_gate}
\end{equation}

Let \(
\epsilon_\theta(s_t,a_t)
=
\Pr_{A\sim\mu(\cdot\mid s_t)}
\left(
C_\theta(s_t,A,a_t)
\neq
\mathbb{I}
\left[
Q^\pi(s_t,A)>Q^\pi(s_t,a_t)
\right]
\right)
\)
denote $C_\theta$'s error under $\mu$. The estimation error then consists of comparator error and finite-candidate sampling error.

\begin{proposition}[Estimation accuracy of the candidate-relative gate]
\label{prop:rho_accuracy}
For any fixed \((s_t,a_t)\), assuming that the \(K\) candidates are sampled either independently from \(\mu(\cdot\mid s_t)\), or uniformly without replacement from a finite candidate pool whose empirical distribution defines \(\mu(\cdot\mid s_t)\): 
\begin{equation}
\left|
\mathbb{E}\!\left[\widehat\rho_{\theta,K}\right]
-
\rho_\mu(s_t,a_t)
\right|
\leq
\epsilon_\theta(s_t,a_t).
\label{eq:rho_bias}
\end{equation}
Moreover, with probability at least \(1-\delta\), we have: 
\begin{equation}
\left|
\widehat\rho_{\theta,K}
-
\rho_\mu(s_t,a_t)
\right|
\leq
\epsilon_\theta(s_t,a_t)
+
\sqrt{
\frac{\log(2/\delta)}{2K}
}.
\label{eq:rho_concentration}
\end{equation}
\end{proposition}

The first term captures errors made by $C_\theta$, while the second is the standard Monte Carlo error from using only \(K\) candidates. Consequently, when the true domination rate $\rho_\mu(s_t,a_t)$ is sufficiently far from the threshold \(\frac{R}{K}\), the learned gate $g_\theta
    =
    \mathbb I\!\left[
        \widehat\rho_{\theta,K}\geq R/K
    \right]$ makes the same intervention decision as the oracle candidate-relative gate $\mathbb I[\rho_\mu(s_t,a_t)\geq R/K]$ with probability at least $1-\delta$. We provide the detailed proof and analysis in Appendix~\ref{app:method_theory}. 
When intervention occurs, the same comparisons that establish that \(a_t\) is weak also identify alternatives predicted to have higher continuation value. \name therefore obtains both components of constructive runtime intervention---\emph{when} to interrupt and \emph{what} direction to expose---from the same comparison primitive.
\section{Experiments}
\label{sec:evaluation}

\subsection{Setups}
\label{sec:evaluation_setup}

\noindent \textbf{Models.}
All agents in our experiments follow a ReAct-style interaction loop that alternates a textual rationale, one environment action, and the resulting observation. 
We evaluate Qwen3-8B~\footnote{https://huggingface.co/Qwen/Qwen3-8B}~\citep{qwen3}, Qwen3.6-35B-A3B~\footnote{https://huggingface.co/Qwen/Qwen3.6-35B-A3B}~\citep{qwen3.6-35b-a3b}, and DeepSeek-V4-Flash~\citep{deepseekai2026deepseekv4}.
Unless stated otherwise, the tiny comparator in \name is a full-parameter fine-tuned
Qwen2.5-0.5B-Instruct~\footnote{https://huggingface.co/Qwen/Qwen2.5-0.5B-Instruct}~\citep{qwen2.5} comparator. The actor
itself remains frozen in every condition.

\noindent \textbf{Benchmarks.} 
We study three environments with complementary action interfaces.
\noindent \text{(1) WebShop} requires multi-step product search and purchasing under
natural-language constraints, and provides partial reward in $[0,1]$
\citep{yao2022webshop}. We collect branch supervision from tasks 0--4999 and
evaluate on the disjoint tasks 5000--5499. 
\noindent \text{(2) ALFWorld} evaluates
compositional household tasks in a text-based embodied environment
\citep{shridhar2020alfworld}; we use its standard training split for data
collection and the valid-unseen split for evaluation. 
\noindent \text{(3) $\tau^3$-Retail}
tests policy-constrained customer-service dialogue with state-changing tools
and a simulated user~\citep{yao2024tau}; we retain the benchmark's official
train/test partition. Splits are always made at the task level.

\noindent \textbf{Metrics.}
WebShop uses mean reward in $[0,1]$ as its primary metric, while ALFWorld and $\tau^3$-Retail use success rate. For ease of comparison, all tables in the main content report performance on a common $0$--$100$ scale: WebShop mean reward is multiplied by $100$, while ALFWorld and $\tau^3$-Retail success rates are reported in percentage points.
This rescaling is purely for presentation and does not change the underlying evaluation metrics.
Qwen actors are evaluated on 500 WebShop tasks and 134 ALFWorld valid-unseen games. Retail uses all 40 test tasks with three seeds, giving 120 episodes per system. To control cost, DeepSeek-V4-Flash uses fixed subsets of the same held-out WebShop and ALFWorld
tasks, while retaining the full three-seed Retail protocol. 
We additionally report estimated token-based inference cost, computed from recorded input and output token counts using OpenRouter public list prices as of August 2026. Detailed pricing and cost accounting are provided in Appendix~\ref{app:api_cost_account_rule}.

\noindent \textbf{Baselines.}
We compare \name against three representative baselines.  
(1)~\emph{Self-Reflection} lets the actor reconsider and revise its own proposal without external feedback~\citep{shinn2023reflexion}.
(2)~\emph{Asym-AC} represents critic-based runtime intervention, where a separate Qwen3-8B critic provides free-form corrective feedback before the actor replans~\citep{jiang2026asymmetric}. 
(3)~\emph{AgentRM} represents reward-model-based inference-time search: it performs beam search over multiple candidate trajectories and selects outputs according to the learned reward model~\citep{xia2025agentrm}. 
Prompt templates in all benchmarks are provided in Appendix~\ref{app:baseline_prompts}. 
We detail the baseline selection and implementations in Appendix~\ref{app:baseline_implementations}.

\subsection{Implementation Details}
\label{sec:implementation}
\noindent\textbf{Prefix-branch supervision.}
\name constructs comparator supervision with respect to a chosen frozen source actor. In our experiments, we use Qwen3-8B as this source actor for all three environments. 
We first complete a base trajectory, then restore selected prefixes and vary
only the first branch action before handing control back to the same
continuation actor. Branches are one level deep and are labeled solely by
downstream environment outcomes. 
The branch sampling policy,
and offline generation cost are deferred to
Appendix~\ref{app:branch_collection} and Appendix~\ref{app:cost_offline_details}. 

\noindent\textbf{Comparator implementation.}
Each example contains the actor-visible state and an ordered action pair, with labels \texttt{A}, \texttt{B}, or \texttt{T} denoting preference for the first action, preference for the second, or a tie. Each physical action pair is included in both A/B orders. After preprocessing, this yields 55k, 23k, and 9k supervision examples for WebShop, ALFWorld, and $\tau^3$-Retail, respectively. At deployment, \name accepts a winner only when the predictions under both input orders are semantically consistent. Full label construction and training details are provided in Appendix~\ref{app:comparator_ABT}. 

\noindent\textbf{Intervention hyperparameters.}
$K$ is the number of alternatives provided each time, and $R$ is the
minimum number that must consistently beat the proposal $a_t$ before intervention. 
We choose \(K\) and \(R\) by grid search on validation tasks using end-to-end task performance. This yields $K{=}4,R{=}1$ on WebShop and ALFWorld and $K{=}4,R{=}2$ on
$\tau^3$-Retail. Details are provided in Appendix~\ref{app:kr_sweep}. 

\noindent\textbf{Reference candidate mechanisms.}
We instantiate \(\mu\) in three ways: sampling legal environment actions, retrieving existing actions from offline data, or generating alternative actions with a 0.5B small language model. WebShop uses sampled environment actions; ALFWorld uses sampled environment actions, augmented with small-LM generation for stronger actors; and \(\tau^3\)-Retail combines small-LM generation with offline retrieval. The comparator and intervention rule are unchanged across these configurations. Full construction details are provided in Appendix~\ref{app:candidate_sources}.

\subsection{Main Results}
\label{sec:evaluation_main}

\begin{table*}[t]
\centering
\caption{
Comparison with baselines.
Performance (Perf.) is mean reward on WebShop
and success rate on ALFWorld and $\tau^3$-Retail. 
Best performance is \textbf{bolded} and the second-highest distinct performance
is \underline{underlined}; the highest inference cost in each actor--environment
setting is shown in \textcolor{costred}{red}.
Cost is the estimated API-equivalent inference cost in USD per 100 tasks.
Avg.~T is the mean task time normalized by the corresponding Original actor.
}
\label{tab:method_comparison}

\resizebox{\textwidth}{!}{%
\begin{tabular}{llccccccccc}
\toprule
& & \multicolumn{3}{c}{WebShop}
& \multicolumn{3}{c}{ALFWorld}
& \multicolumn{3}{c}{$\tau^3$-Retail} \\
\cmidrule(lr){3-5}
\cmidrule(lr){6-8}
\cmidrule(lr){9-11}
Actor & Method
& Perf.$\uparrow$ & Cost$\downarrow$ & Avg.~T$\downarrow$
& Perf.$\uparrow$ & Cost$\downarrow$ & Avg.~T$\downarrow$
& Perf.$\uparrow$ & Cost$\downarrow$ & Avg.~T$\downarrow$ \\
\midrule

\multirow{5}{*}{Qwen3-8B}
& Original
& 39.6 & \$0.698 & 1.00$\times$
& 82.8 & \$0.274 & 1.00$\times$
& 37.5 & \$2.00 & 1.00$\times$ \\

& Self-Reflection
& 35.3 & \$1.54 & 2.45$\times$
& 83.6 & \$0.630 & 1.03$\times$
& 39.2 & \$4.42 & 1.66$\times$ \\

& Asym-AC
& 10.1 & \$1.15 & 2.27$\times$
& 82.8 & \$0.492 & 1.03$\times$
& 39.2 & \$3.20 & 2.30$\times$ \\

& AgentRM
& \underline{40.1} & \textcolor{costred}{\$12.2} & 16.8$\times$
& \underline{85.1} & \textcolor{costred}{\$3.61} & 20.6$\times$
& \textbf{50.8} & \textcolor{costred}{\$32.8} & 21.6$\times$ \\

& \name (ours)
& \textbf{56.3} & \$0.883 & 1.41$\times$
& \textbf{90.3} & \$0.378 & 1.14$\times$
& \underline{45.0} & \$2.19 & 2.04$\times$ \\

\midrule

\multirow{5}{*}{Qwen3.6-35B-A3B}
& Original
& 56.6 & \$0.362 & 1.00$\times$
& 85.1 & \$0.218 & 1.00$\times$
& 62.5 & \$0.613 & 1.00$\times$ \\

& Self-Reflection
& 58.1 & \$0.736 & 2.00$\times$
& 85.1 & \$0.475 & 1.57$\times$
& 62.5 & \$1.55 & 2.08$\times$ \\

& Asym-AC
& 41.6 & \$0.666 & 2.22$\times$
& 85.8 & \$0.438 & 1.43$\times$
& 60.8 & \$0.911 & 1.42$\times$ \\

& AgentRM
& \underline{62.4} & \textcolor{costred}{\$5.83} & 9.19$\times$
& \underline{87.3} & \textcolor{costred}{\$3.23} & 9.22$\times$
& \textbf{68.8} & \textcolor{costred}{\$12.0} & 9.65$\times$ \\

& \name (ours)
& \textbf{68.1} & \$0.636 & 1.45$\times$
& \textbf{94.0} & \$0.231 & 1.30$\times$
& \underline{65.0} & \$0.670 & 1.78$\times$ \\

\midrule

\multirow{5}{*}{DeepSeek-V4-Flash}
& Original
& 60.9 & \$8.51 & 1.00$\times$
& 90.0 & \$0.133 & 1.00$\times$
& 80.8 & \$0.825 & 1.00$\times$ \\

& Self-Reflection
& 78.0 & \$6.99 & 0.73$\times$
& \underline{95.0} & \$0.204 & 0.91$\times$
& 77.5 & \$1.40 & 1.13$\times$ \\

& Asym-AC
& 70.3 & \$6.26 & 0.58$\times$
& 87.5 & \$0.268 & 0.94$\times$
& 80.0 & \$0.897 & 0.79$\times$ \\

& AgentRM
& \textbf{81.7} & \textcolor{costred}{\$76.4} & 8.81$\times$
& \textbf{100} & \textcolor{costred}{\$0.645} & 8.71$\times$
& \textbf{87.5} & \textcolor{costred}{\$8.45} & 11.0$\times$ \\

& \name (ours)
& \underline{79.6} & \$3.19 & 0.49$\times$
& \textbf{100} & \$0.083 & 0.76$\times$
& \underline{82.5} & \$0.882 & 1.17$\times$ \\

\bottomrule
\end{tabular}%
}
\end{table*}

\noindent\textbf{Performance.}
Table~\ref{tab:method_comparison} shows that \name consistently improves the original actor and achieves either the best or second-best performance in every setting.
For Qwen3-8B, performance increases from $0.396$ to $0.563$ on WebShop, from $82.8\%$ to $90.3\%$ on ALFWorld, and from $37.5\%$ to $45.0\%$ on $\tau^3$-Retail.
The gains persist for the substantially stronger Qwen3.6-35B-A3B and DeepSeek-V4-Flash.
This is important because the same benchmark-specific 0.5B comparator is not merely compensating for a weak actor: it remains useful even when the proposal is produced by substantially stronger models. 
More notably, each benchmark-specific comparator is trained only from Qwen3-8B branch supervision and reused unchanged for the stronger actors. Our cross-actor analysis in Appendix~\ref{app:cross_actor} shows that source-actor preferences predominantly attenuate to ties rather than reverse direction under stronger continuation actors, providing a possible explanation for this transfer. 

\name outperforms both Self-Reflection and Asym-AC in all nine settings.
AgentRM provides a substantially stronger comparison: its beam-search inference matches or exceeds \name in several settings, particularly on $\tau^3$-Retail and with DeepSeek-V4-Flash.
However, this additional performance comes from substantially more inference-time search.
Overall, \name provides consistent gains across actor scales while remaining competitive with much more expensive search-based inference.

\noindent\textbf{Inference cost.}
\name achieves these gains with relatively small additional token cost.
Its API-equivalent cost is higher than the Original actor in most settings, as expected from the ``replanning design”, but remains lower than both Self-Reflection and Asym-AC in all nine settings while simultaneously achieving higher performance.
The contrast with AgentRM is larger: AgentRM costs $7.8$--$24.0\times$ more than \name across the nine settings.
Thus, although search can occasionally attain higher performance, \name provides a substantially cheaper way to improve the actor without requiring a large inference-time search budget.

\noindent\textbf{Runtime.}
\name incurs modest online overhead, averaging $1.28\times$ the end-to-end episode time of the original actor, with seven of nine settings below $1.5\times$. The larger overheads occur on $\tau^3$-Retail, where the chosen reference candidate mechanism additionally invokes a small LLM. By comparison, AgentRM requires $5.4$--$18.1\times$ as much episode time as \name because of its multi-trajectory beam search. 
Avg.~T can occasionally fall below $1\times$ because useful intervention may reduce the number of interaction steps required to complete the task. 

\subsection{Ablation Study}
\label{sec:evaluation_ablation}

Holding $\mu$ fixed, we ablate two design choices in \name: the \emph{learning objective}, comparing absolute $Q^\pi(s,a)$ estimation with our pairwise comparator $C_\theta$, and the \emph{intervention mechanism}, comparing forced action execution with constructive runtime intervention as defined in Section~\ref{sec:problem}. This yields a $2\times2$ ablation, which we evaluate with Qwen3-8B and Qwen3.6-35B-A3B. In this ablation, the absolute-\(Q\) estimation model is trained by fine-tuning the same Qwen2.5-0.5B backbone.

\begin{table*}[t]
\centering
\caption{Ablation of learning objective and intervention mechanism. WebShop reports mean reward; ALFWorld and $\tau^3$-Retail report success rate. ``Constructive'' denotes constructive runtime intervention: the proposed action $a_t$ is withheld only when intervention is triggered, a preferred alternative is returned as advice, and the actor $\pi$ replans before execution.}
\label{tab:q_control_ablation}
\small
\begin{tabular}{llrrrrrr}
\toprule
& & \multicolumn{2}{c}{WebShop} & \multicolumn{2}{c}{ALFWorld}
& \multicolumn{2}{c}{$\tau^3$-Retail} \\
\cmidrule(lr){3-4}\cmidrule(lr){5-6}\cmidrule(lr){7-8}
Objective & Intervention & Qwen3 & Qwen3.6 & Qwen3 & Qwen3.6 & Qwen3 & Qwen3.6 \\
\midrule
Absolute $Q$ & Forced & 31.5 & 41.9 & 2.24 & 8.96 & 4.17 & 5.00 \\
Absolute $Q$ & Constructive & 54.9 & 64.5 & 57.5 & 63.4 & 16.7 & 17.5 \\
Pairwise comparison & Forced & 37.8 & 60.4 & 51.5 & 50.8 & 16.7 & 37.5 \\
Pairwise comparison & Constructive & \textbf{56.3} & \textbf{68.1}
& \textbf{90.3} & \textbf{94.0} & \textbf{45.0} & \textbf{65.0} \\
\bottomrule
\end{tabular}
\end{table*}

Table~\ref{tab:q_control_ablation} shows that both design choices are important. First, directly executing the selected alternative action is consistently inferior to constructive intervention. Under absolute-$Q$ supervision, replacing forced execution with constructive intervention raises Qwen3-8B from $2.24\%$ to $57.46\%$ on ALFWorld and from $4.17\%$ to $16.67\%$ on $\tau^3$-Retail, with the same trend for Qwen3.6. The pattern remains under pairwise supervision: forcing the predicted winner is substantially worse than returning it as advice and allowing the actor to replan, particularly on the two long-horizon environments. These results support the formulation in Section~\ref{sec:problem}: a useful advisor should guide the stronger actor rather than directly take over its next action.

Constructive intervention alone, however, does not account for the full gain. Holding $\mu$ fixed, replacing absolute-$Q$ estimation with same-prefix pairwise comparison improves ALFWorld from $57.46\%$ to $90.30\%$ for Qwen3-8B and from $63.43\%$ to $94.03\%$ for Qwen3.6; on $\tau^3$-Retail, the corresponding gains are $16.67\%\rightarrow45.00\%$ and $17.50\%\rightarrow65.00\%$. 
The pairwise objective directly matches the judgment required within \name—whether an available alternative is better than current $a_t$—without requiring the comparator to estimate an absolute long-horizon value.
The best performance therefore requires both components of \name: a narrow comparison objective and constructive runtime intervention that delegates final replanning to the actor.

\subsection{Constructive Replanning}
\label{exp:actor_control}

\begin{wrapfigure}{r}{0.6\textwidth}
\centering
\includegraphics[width=\linewidth]{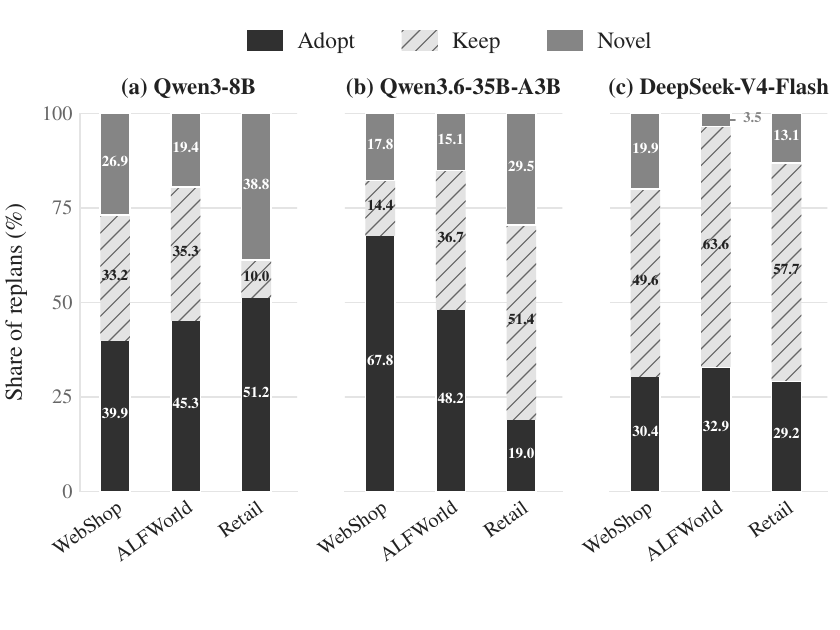}
\caption{
\textbf{Actor behavior after \name intervention.}
The actor's replan decision $a_t'$ are categorized as
\emph{Adopt} if they adopt the advice $a_t^{adv}$,
\emph{Keep} if they repeat the original $a_t$,
and \emph{Novel} otherwise.
}
\label{fig:actor_replan}
\end{wrapfigure}

A central property of \name is that intervention provides additional local
directions without replacing the actor's own decision process. We first ask
whether the reference candidate mechanism $\mu$ exposes actions that repeated actor
sampling would otherwise miss. Across 300 prefixes, eight high-temperature
actor samples produce only 2.28 distinct executable actions on average
(28.5\% unique-action utilization), with all eight samples collapsing to the
same canonical action for 150 prefixes. In contrast, four reference candidates
produce 3.71 distinct actions on average (92.7\% utilization), showing that the
reference mechanism substantially broadens the local action support available
for intervention.

We then examine how this additional information affects replanning.
Figure~\ref{fig:actor_replan} categorizes the actor's next proposal as
\emph{Adopt}, \emph{Keep}, or \emph{Novel}. Actors frequently adopt recommended
actions, but substantial mass remains on \emph{Keep} and \emph{Novel},
confirming that advice does not reduce replanning to forced selection.
For example, DeepSeek-V4-Flash adopts only 29.2--32.9\% of recommendations
across the three environments while retaining its original proposal in
49.6--63.6\% of replans. Qwen3.6 is more context-dependent, adopting
67.8\% on WebShop but only 19.0\% on $\tau^3$-Retail. These results show that
\name can expose otherwise-missing local directions while preserving the
actor's control over the final action.

\subsection{Role of Repeated Intervention}
\label{exp:why_COTA_works}
\begin{table}[t]
\centering
\small
\caption{
Performance under different episode-level intervention budgets with Qwen3-8B as the actor. \(B\) is the maximum number of decision states at which \name may intervene within an episode; “Full” removes this cap. The “$>2$ interv.” column reports the fraction of Full-\name episodes with more than two interventions.
}
\label{tab:intervention_budget}
\begin{tabular}{lccccc}
\toprule
Environment & $B=0$ & $B=1$ & $B=2$ & Full & $>2$ interv. \\
\midrule
WebShop   & 36.6 & 40.6 & 44.1 & \textbf{56.3} & 68.0\% \\
ALFWorld  & 82.8 & 83.6 & 79.9 & \textbf{90.3} & 37.3\% \\
$\tau^3$-Retail & 44.2 & 43.3 & 44.2 & \textbf{45.0} & 19.2\% \\
\bottomrule
\end{tabular}
\end{table}

Unlike one-shot correction, \name remains active throughout the trajectory and may intervene again after the actor replans. We therefore ask whether its gains come from a few isolated interventions or from keeping the intervention loop active over the episode. To test this, we cap the number of decision states at which \name may intervene by an episode-level budget $B$. 

We use Qwen3-8B as the actor. As shown in Table~\ref{tab:intervention_budget}, restricting \name to one or two interventions leaves substantial performance unrealized on WebShop and ALFWorld. On WebShop, mean reward increases from $0.366$ at $B=0$ to $0.406$ and $0.441$ at $B=1$ and $B=2$, but reaches $0.563$ with unrestricted intervention; correspondingly, 68.0\% of Full-\name episodes contain more than two interventions. ALFWorld shows the same overall pattern despite non-monotonic small-budget results: $B=1$ and $B=2$ reach only $83.58\%$ and $79.85\%$, compared with $90.30\%$ under Full-\name. In contrast, $\tau^3$-Retail is relatively insensitive to the budget, consistent with only 19.2\% of episodes containing more than two interventions. These results suggest that \name's benefit is not necessarily concentrated in a single decisive correction; when useful intervention opportunities recur, keeping the lightweight comparison loop active throughout the trajectory can be important.

\section{Related Work}
\label{sec:related_work}

\noindent\textbf{Runtime Intervention.}
Runtime intervention has roots in human intervention and shielding for sequential decision making, where an external mechanism blocks or modifies unsafe actions~\citep{saunders2018trial,alshiekh2018shielding}. 
Recent LLM-agent guardrails extend this idea through policy enforcement and proactive risk monitoring~\citep{chen2025shieldagent,wang2026agentspec,wang2026probguard}. 
Beyond safety, \citet{vasudev2026accurate} show that accurate failure prediction does not necessarily yield beneficial intervention, while \citet{zhang2026calibration} argue that runtime control should account for the downstream advantage of available interventions. 
Building on this shift from detecting failures to evaluating intervention utility, we study whether the mechanism that determines and guides intervention can itself be made lightweight.

\noindent\textbf{Lightweight Auxiliary Models for Strong Actors.}
A growing line of work shows that smaller specialized models can assist substantially stronger executors through verification, workflow optimization, or critique~\citep{kirchner2024prover,nie2025weak,dong2025rag,yang2025lighthouse}. 
In agent settings, \citet{jiang2026asymmetric} use a weaker critic for runtime intervention, while \citet{jin2026weak} show that useful revision guidance can improve a stronger frozen model without requiring the critic to solve the task end-to-end. 
Building on this separation of roles, we push the idea further by narrowing the capability required of the auxiliary model from task-level solving or generation to local pairwise comparison, making the intervention role viable for a truly lightweight model.

\section{Conclusion}
\label{sec:conclusion}

We study how to improve long-horizon LLM agents without further post-training the actor or relying on another strong but expensive LLM for runtime intervention.
Our results show that the intervention role can be separated from task solving and specialized into a lightweight module: \name trains a 0.5B comparator for local action comparison while leaving task-level replanning to the original actor.
Across diverse environments and actors, this design consistently improves the underlying agent while maintaining a favorable performance--cost trade-off.
More broadly, our results suggest a practical division of labor for agent systems: keep strong LLMs focused on solving the task, and delegate narrow intervention decisions to cheap, specialized models.

\bibliography{smalladvisors_references}
\bibliographystyle{iclr2026_conference}

\appendix

\section{Theoretical Details}
\label{app:method_theory}

This section provides additional analysis for the methodology in
Section~\ref{sec:method}. 

\subsection{Same-Prefix Branch Analysis}
\label{app:ta_same_prefix}

Fix a decision state $s_t$, a branch-point action $a$, and the frozen
continuation actor $\pi$. By construction, a same-prefix branch first restores
the environment to $s_t$, executes $a$, and then returns control to the same
actor $\pi$. Recall that: 
\begin{equation}
    Y(s_t,a)
    =
    R(\tau),
    \qquad
    \tau
    \sim
    P(\cdot\mid s_t,a,\pi).
    \label{eq:app_ta_branch_return}
\end{equation}
Therefore,
\begin{align}
    \mathbb E[Y(s_t,a)]
    &=
    \mathbb E_{\tau\sim P(\cdot\mid s_t,a,\pi)}
    [R(\tau)]
    \nonumber\\
    &=
    Q^\pi(s_t,a),
    \label{eq:app_ta_branch_unbiased}
\end{align}
which gives Eq.~\ref{eq:branch_expectation} in the main text.

For $M$ independent continuations,
\begin{equation}
    \widehat Q_M(s_t,a)
    =
    \frac{1}{M}
    \sum_{m=1}^{M}
    Y_m(s_t,a),
    \label{eq:app_ta_mc_value}
\end{equation}
and linearity of expectation gives: 
\begin{equation}
    \mathbb E[
        \widehat Q_M(s_t,a)
    ]
    =
    Q^\pi(s_t,a).
    \label{eq:app_ta_mc_unbiased}
\end{equation}
Whenever the return is integrable,
$\widehat Q_M(s_t,a)$ also converges almost surely to
$Q^\pi(s_t,a)$ as $M\rightarrow\infty$.

For two sibling actions $a_A$ and $a_B$, define their true and empirical
continuation-value gaps as: 
\begin{align}
    d(s_t,a_A,a_B)
    &=
    Q^\pi(s_t,a_A)
    -
    Q^\pi(s_t,a_B),
    \label{eq:app_ta_true_gap}
    \\
    \widehat d_M(s_t,a_A,a_B)
    &=
    \widehat Q_M(s_t,a_A)
    -
    \widehat Q_M(s_t,a_B).
    \label{eq:app_ta_empirical_gap}
\end{align}
Then we have: 
\begin{equation}
    \mathbb E[
        \widehat d_M(s_t,a_A,a_B)
    ]
    =
    d(s_t,a_A,a_B).
    \label{eq:app_ta_gap_unbiased}
\end{equation}

Thus the branch-return difference isolates the downstream consequence of
changing the branch-point action while keeping both the preceding state and
the continuation actor fixed. This is the quantity thresholded to construct
the empirical supervision in Eq.~\ref{eq:pairwise_label}.

It is useful to distinguish unbiased value estimation from finite-sample
pairwise labeling. Although
$\widehat Q_M(s_t,a)$ is unbiased for $Q^\pi(s_t,a)$, the thresholded label
$\mathbb I[\widehat d_M>0]$ need not be an unbiased estimator of
$\mathbb I[d>0]$. In particular, comparisons with small true value gaps are
more susceptible to finite-rollout noise.

For example, suppose branch returns are normalized to $[0,1]$, and suppose
$\widehat d_M$ is formed from $M$ independent paired branch-return
differences, each lying in $[-1,1]$. If $d>0$, Hoeffding's inequality gives: 
\begin{equation}
    \Pr(
        \widehat d_M
        \leq
        0
    )
    \leq
    \exp\!\left(
        -\frac{M d^2}{2}
    \right).
    \label{eq:app_ta_pairwise_label_error}
\end{equation}
Hence increasing the number of branch continuations primarily improves the
reliability of supervision for comparisons whose true continuation values are
close.

\subsection{Proof of Proposition~\ref{prop:rho_accuracy}}
\label{app:ta_rho_proof}

Fix $(s_t,a_t)$ throughout this subsection. For a candidate
$A\sim\mu(\cdot\mid s_t)$, define: 
\begin{align}
    B(A)
    &=
    \mathbb I
    \left[
        Q^\pi(s_t,A)
        >
        Q^\pi(s_t,a_t)
    \right],
    \label{eq:app_ta_oracle_pair}
    \\
    \widetilde B(A)
    &=
    C_\theta(s_t,A,a_t).
    \label{eq:app_ta_predicted_pair}
\end{align}
For compactness, write: 
\begin{align}
    \rho
    &=
    \rho_\mu(s_t,a_t)
    =
    \mathbb E_{A\sim\mu}[B(A)],
    \label{eq:app_ta_rho_short}
    \\
    q_\theta
    &=
    \mathbb E_{A\sim\mu}[\widetilde B(A)].
    \label{eq:app_ta_qtheta}
\end{align}
The pairwise error defined in Section~\ref{sec:gate_accuracy} can be written as: 
\begin{equation}
    \epsilon_\theta
    =
    \Pr_{A\sim\mu}
    \left(
        \widetilde B(A)
        \neq
        B(A)
    \right).
    \label{eq:app_ta_pairwise_error_short}
\end{equation}

Because both $B(A)$ and $\widetilde B(A)$ are binary, we have: 
\begin{align}
    |q_\theta-\rho|
    &=
    \left|
        \mathbb E[
            \widetilde B(A)-B(A)
        ]
    \right|
    \nonumber\\
    &\leq
    \mathbb E[
        |\widetilde B(A)-B(A)|
    ]
    \nonumber\\
    &=
    \epsilon_\theta.
    \label{eq:app_ta_bias_core}
\end{align}
This step does not require independence across candidates; it is purely a
single-candidate comparison between the learned and oracle relations.

Now suppose first that
$A_1,\ldots,A_K$ are independently sampled from
$\mu(\cdot\mid s_t)$, and define: 
\begin{equation}
    \widetilde B_i
    =
    \widetilde B(A_i),
    \qquad
    \widehat\rho_{\theta,K}
    =
    \frac{1}{K}
    \sum_{i=1}^{K}
    \widetilde B_i.
    \label{eq:app_ta_empirical_learned_rho}
\end{equation}
By linearity of expectation, we have: 
\begin{equation}
    \mathbb E[
        \widehat\rho_{\theta,K}
    ]
    =
    q_\theta.
    \label{eq:app_ta_empirical_rho_expectation}
\end{equation}
Combining
Eqs.~\eqref{eq:app_ta_bias_core} and
\eqref{eq:app_ta_empirical_rho_expectation} yields: 
\begin{equation}
    \left|
        \mathbb E[
            \widehat\rho_{\theta,K}
        ]
        -
        \rho
    \right|
    \leq
    \epsilon_\theta,
    \label{eq:app_ta_bias_result}
\end{equation}
which proves Eq.~\ref{eq:rho_bias}.

For the finite-candidate component, 
for independent candidate sampling, the concentration result follows directly
from Hoeffding's inequality. The same bound also holds when the $K$ candidates
are sampled uniformly without replacement from a finite candidate population:
sampling without replacement is at least as concentrated as sampling with
replacement for bounded finite-population averages. Therefore, under either
sampling scheme, we have: 

\begin{equation}
    \Pr\!\left(
        \left|
            \widehat\rho_{\theta,K}
            -
            q_\theta
        \right|
        \geq
        \xi
    \right)
    \leq
    2\exp(-2K\xi^2).
    \label{eq:app_ta_iid_hoeffding}
\end{equation}
With probability at least $1-\delta$,
\begin{equation}
    \left|
        \widehat\rho_{\theta,K}
        -
        q_\theta
    \right|
    \leq
        \sqrt{
        \frac{\log(2/\delta)}{2K}
    }.
    \label{eq:app_ta_iid_concentration}
\end{equation}
Using the triangle inequality together with
Eq.~\ref{eq:app_ta_bias_core}, we have: 
\begin{align}
    \left|
        \widehat\rho_{\theta,K}
        -
        \rho
    \right|
    &\leq
    \left|
        \widehat\rho_{\theta,K}
        -
        q_\theta
    \right|
    +
    |q_\theta-\rho|
    \nonumber\\
    &\leq
    \epsilon_\theta
    +
    \sqrt{
        \frac{\log(2/\delta)}{2K}
    }.
    \label{eq:app_ta_iid_total_error}
\end{align}
This proves Eq.~\ref{eq:rho_concentration} under both sampling schemes.

\subsection{Gate Decision Consistency}
\label{app:ta_gate_agreement}

Proposition~\ref{prop:rho_accuracy} implies that, with probability at least
$1-\delta$: 
\begin{equation}
    \left|
        \widehat\rho_{\theta,K}
        -
        \rho_\mu(s_t,a_t)
    \right|
    \leq
    \epsilon_\theta(s_t,a_t)
    +
    \sqrt{
        \frac{\log(2/\delta)}{2K}
    }.
    \label{eq:app_ta_gate_confidence_event}
\end{equation}

Therefore, whenever the true domination rate is separated from the
intervention threshold by more than this error: 
\begin{equation}
    \left|
        \rho_\mu(s_t,a_t)
        -
        \frac{R}{K}
    \right|
    >
    \epsilon_\theta(s_t,a_t)
    +
    \sqrt{
        \frac{\log(2/\delta)}{2K}
    },
    \label{eq:app_ta_gate_margin}
\end{equation}
the learned estimate and the true domination rate lie on the same side of the
threshold with probability at least $1-\delta$. Consequently, we have: 
\begin{equation}
    \mathbb I
    \left[
        \widehat\rho_{\theta,K}
        \geq
        \frac{R}{K}
    \right]
    =
    \mathbb I
    \left[
        \rho_\mu(s_t,a_t)
        \geq
        \frac{R}{K}
    \right]
    \label{eq:app_ta_gate_decision_agreement}
\end{equation}
with probability at least $1-\delta$.

Thus, away from an uncertainty region around the intervention threshold, the
learned gate agrees with the oracle candidate-relative gate. Lower comparator
error narrows this region through $\epsilon_\theta$, while increasing $K$
reduces the finite-candidate sampling term at the standard $K^{-1/2}$ rate.

\subsection{Candidate Distribution}
\label{app:ta_candidate_support}

The candidate-relative formulation does not require the reference mechanism
$\mu$ to be an expert policy. However, constructive advice is necessarily
limited by candidate support: the framework cannot expose a genuinely better
alternative if the sampled candidate set contains none.

Under independent sampling from $\mu$, the probability that at least one of
$K$ candidates truly improves on the current proposal is: 
\begin{equation}
    \Pr\!\left[
        \exists i:
        Q^\pi(s_t,A_i)
        >
        Q^\pi(s_t,a_t)
    \right]
    =
    1-
    \left[
        1-\rho_\mu(s_t,a_t)
    \right]^K.
    \label{eq:app_ta_support_iid}
\end{equation}

For the finite-pool without-replacement setting, suppose exactly $W$ of the $N$
available candidates have higher continuation value than the proposal. Then: 
\begin{equation}
    \rho_N
    =
    \frac{W}{N}.
    \label{eq:app_ta_support_fraction}
\end{equation}
The probability that a uniformly sampled subset of size $K$ contains at least
one true winner is: 
\begin{equation}
    \Pr(
        \text{at least one true winner}
    )
    =
    1-
    \frac{
        \binom{N-W}{K}
    }{
        \binom{N}{K}
    },
    \label{eq:app_ta_support_wor}
\end{equation}
where the numerator is interpreted as zero when $K>N-W$.

These expressions describe a support ceiling rather than a quality assumption
on $\mu$. A weak reference distribution can still provide a meaningful
candidate-relative comparison, but it will supply genuinely useful
alternatives less frequently. Conversely, broader or stronger candidate
support increases the opportunities for constructive replanning without
changing the comparator's learning objective. 
Appendix~\ref{app:candidate_support} provides a controlled empirical study of this support effect on WebShop. 

\section{Experimental Details}
\label{app:experiments}

\subsection{Evaluation Protocols and Data Splits}
\label{app:actor_protocols}

\begin{table}[H]
\centering
\caption{Evaluation slices. A repeat denotes one complete environment episode
with a distinct seeded rollout. DeepSeek uses cost-controlled fixed subsets for
WebShop and ALFWorld.}
\label{tab:evaluation_slices}
\small
\resizebox{\columnwidth}{!}{%
\begin{tabular}{lrrrrr}
\toprule
Environment & Local tasks & DeepSeek tasks & Local repeats & DeepSeek repeats & Max steps \\
\midrule
WebShop & 500 & 50 & 1 & 1 & 20 \\
ALFWorld & 134 & 20 & 1 & 2 & 30 \\
$\tau^3$-Retail & 40 & 40 & 3 & 3 & 100 \\
\bottomrule
\end{tabular}%
}
\end{table}

\noindent \textbf{WebShop.}
The branch-data pool uses tasks 0--4999. The held-out evaluation set is tasks
5000--5499, which is never used for branching, preprocessing, checkpoint
selection, or hyperparameter sweeps. The environment returns a reward in
$[0,1]$ based on matched product attributes. We call reward 1 an exact success
and separately report the fraction of reward-0 trajectories.

\noindent \textbf{ALFWorld.}
Data collection uses training games, while evaluation uses the 134 games in
the standard valid-unseen split. A trajectory succeeds only when the
environment's terminal goal condition is satisfied. The actor sees the task,
the executed action--observation prefix, the current observation, and the
current admissible action set.

\noindent \textbf{$\tau^3$-Retail.}
We use the official Retail test split of 40 tasks. Each system is evaluated
under three rollout seeds, and success is averaged over all 120 episodes. The
environment includes a simulated user, policy text, and a database-backed tool
interface; one actor turn may be followed by a tool result or a user reply.

\subsection{Prefix-Branch Collection}
\label{app:branch_collection}

We generate supervision in two stages. First, the frozen Qwen3-8B source actor completes
a linear base trajectory without intervention. Second, after termination, we
select decision points across the trajectory and restore each exact prefix.
Where the environment does not support snapshots, restoration resets the task
and replays all preceding actions. 

For a base trajectory with $T$ actions, the default sampler chooses
$\min(\lceil0.5T\rceil,10)$ temporally stratified decision points. At each
verified point, we keep the base action and obtain up to five distinct legal
alternatives. The branching action differs across siblings; every branch thereafter
returns to the same frozen continuation actor, low-temperature decoding,
remaining step budget, and terminal rule. Branch trajectories are never
branched recursively. This produces linear rather than exponential data growth
while preserving the same-prefix counterfactual interpretation.

\subsection{Training Data Construction}
\label{app:pair_construction}

Within each verified prefix group, we construct both base--branch and
branch--branch pairs. WebShop orders actions by downstream partial reward,
while ALFWorld and Retail use terminal success. Small outcome differences are
treated as ties when they do not support a reliable preference. Ties are
downsampled to approximately 20\% of the training data so that abundant
same-outcome siblings do not dominate learning. Every retained physical pair
is direction-augmented into both $(A,B)$ and $(B,A)$ orders. All members of a
task or interaction tree remain in one split.

\begin{table}[H]
\centering
\caption{Comparator data split details.}
\label{tab:data_details}
\small
\begin{tabular}{lrrrr}
\toprule
Environment & Physical pairs & Train & Validation & Test \\
\midrule
WebShop & 35,117 & 55,856 & 7,078 & 7,300 \\
ALFWorld & 14,548 & 22,678 & 3,222 & 3,196 \\
$\tau^3$-Retail & 5097 & 8,910 & 1284 & 1278 \\
\bottomrule
\end{tabular}
\end{table}

\subsection{Comparator Training and Inference}
\label{app:comparator_ABT}

Section~\ref{sec:tiny_advisor_pipeline} defines the ideal binary comparison
target as $C_\theta(s_t,a_A,a_B)
\approx
\mathbb I
\left[
Q^\pi(s_t,a_A)
>
Q^\pi(s_t,a_B)
\right]$ in Eq.~\ref{eq:comparator_target}.
Here we describe how this oracle target comparator is realized in our implementation. 
Training uses same-prefix branch estimates to supervise the
relative ordering of two actions, with an additional tie class for
indistinguishable outcomes. At inference time, the resulting ternary model
outputs are mapped back to the binary decision \(C_\theta\) required by the
runtime intervention rule.

\paragraph{Training supervision.}
As described in Section~\ref{sec:comparator_training}, for two sibling actions
\(a_A\) and \(a_B\) taken from the same state, we estimate their
actor-conditioned continuation values using same-prefix Monte Carlo branches $\widehat Q_M(s_t,a_A)$ and $\widehat Q_M(s_t,a_B)$. 
The binary relation introduced in Eq.~\ref{eq:pairwise_label} 
$\widehat B(s_t,a_A,a_B)
=
\mathbb I
\left[
\widehat Q_M(s_t,a_A)
>
\widehat Q_M(s_t,a_B)
\right]$ 
is therefore the empirical counterpart of the strict ordering in
Eq.~\ref{eq:comparator_target}. Since
\(\widehat Q_M(s_t,a)\) is a Monte Carlo estimate of \(Q^\pi(s_t,a)\),
the empirical ordering approaches the corresponding ordering of
\(Q^\pi\) as the number of branch continuations increases.

Our implementation uses the same ordering signal, but does not force every
pair into a binary preference. Instead, we train the model with three possible
outputs:
\[
\begin{cases}
\texttt{A},
&
\widehat Q_M(s_t,a_A)
-
\widehat Q_M(s_t,a_B)
>
\gamma_e,
\\[2mm]
\texttt{B},
&
\widehat Q_M(s_t,a_B)
-
\widehat Q_M(s_t,a_A)
>
\gamma_e,
\\[2mm]
\texttt{T},
&
\left|
\widehat Q_M(s_t,a_A)
-
\widehat Q_M(s_t,a_B)
\right|
\leq
\gamma_e.
\end{cases}
\]
Thus, \texttt{A} and \texttt{B} represent the two directions of the same
pairwise ordering used in Eq.~\ref{eq:pairwise_label}, 
while \texttt{T} represents an empirical tie region in the supervision; at deployment, such a prediction contributes no decisive winner. 
When \(\gamma_e=0\), this supervision directly reflects the empirical strict
ordering: unequal branch estimates produce a preference for the action with
the larger estimated continuation value, while equal estimates are labeled
\texttt{T}. For ALFWorld and \(\tau^3\)-Retail, branch outcomes are binary, so
we use \(\gamma_e=0\): outcomes \(1\) versus \(0\) produce a decisive
preference, whereas \(1\) versus \(1\) and \(0\) versus \(0\) produce a tie. 
For WebShop, whose returns provide graded numerical feedback, we use a
positive margin \(\gamma_e=\gamma\). This deliberately treats small empirical
return differences as ties rather than forcing the model to learn a preference
from a weak distinction. The margin is therefore a conservative
implementation-level surrogate for the strict target in
Eq.~\ref{eq:comparator_target}: sufficiently separated action pairs are
trained according to their empirical ordering, whereas pairs within the
margin contribute no decisive preference.

This construction connects directly to the target in the main method.
Because 
$\widehat Q_M(s_t,a)
\rightarrow
Q^\pi(s_t,a)$
as the number of independent branch continuations increases, the training
signal increasingly reflects the ordering of the true actor-conditioned
continuation values. With \(\gamma_e=0\), it approaches the strict relation
in Eq.~\ref{eq:comparator_target} away from exact ties. With
\(\gamma_e>0\), the implementation intentionally abstains on sufficiently
small value differences rather than attempting to resolve them.

\paragraph{Binary comparator at inference.}
Although the model is trained with the three outputs
\(\{\texttt{A},\texttt{B},\texttt{T}\}\), the runtime gate in
Section~\ref{sec:tiny_advisor_pipeline} requires only the binary judgment in
Eq.~\ref{eq:comparator_target}. We obtain this judgment by evaluating
each action pair in both input orders.

Specifically, we set: 
\begin{equation}
C_\theta(s_t,a_A,a_B)
=
\begin{cases}
1,
&
\begin{array}{l}
\text{the model predicts \texttt{A} for }(s_t,a_A,a_B),\\
\text{and predicts \texttt{B} for }(s_t,a_B,a_A),
\end{array}
\\[3mm]
0,
&
\text{otherwise}.
\end{cases}
\label{eq:app_binary_inference}
\end{equation}
Thus, \(a_A\) is counted as defeating \(a_B\) only when both input orders
express the same semantic preference for \(a_A\). A tie prediction, or a pair
of predictions that is inconsistent under input reversal, produces
\(C_\theta=0\). 
Eq.~\ref{eq:app_binary_inference} is the concrete implementation of
the binary \(C_\theta\) used in the main method. In particular, the runtime
prediction in Eq.~\ref{eq:pairwise_decision} is exactly
\(
\widetilde B_i
=
C_\theta(s_t,A_i,a_t),
\)
after applying the bidirectional check above. The winner count in
Eq.~\ref{eq:runtime_reject_rule} and the candidate-relative estimate in
Eq.~\ref{eq:learned_gate} therefore operate on the same binary comparator
defined in Section~\ref{sec:tiny_advisor_pipeline}; the ternary outputs are
only an implementation detail used to obtain that decision conservatively.

We study the empirical effect of ternary supervision and bidirectional
inference in Appendix~\ref{appendix:ABVSABT-exp}.

\subsection{Selection of $K$ and $R$}
\label{app:kr_sweep}

$K$ controls the number of alternatives reviewed at a state and $R$ controls
how many stable winners are required to reject the proposal. We select $K$ and $R$ using end-to-end performance on held-out validation tasks and freeze the selected values before test evaluation. We use the same $K$ and $R$ across actor models within each environment.

\begin{table}[H]
\centering
\caption{Validation performance under different settings of \(K\) and \(R\) on WebShop.}
\label{tab:ablation_system}
\small
\begin{tabular}{lcccc}
\toprule
Configuration & $K$ & $R$ & Mean reward & Success (\%) \\
\midrule
\name, cautious & 4 & 4 & 0.441 & 18.2 \\
\name, medium & 8 & 2 & 0.555 & 28.4 \\
\name, active & 4 & 1 & \textbf{0.635} & \textbf{34.2} \\
\bottomrule
\end{tabular}
\end{table}

After our grid search on respective held-out validation tasks, we select $K{=}4,R{=}1$ for WebShop and
ALFWorld, and $K{=}4,R{=}2$ for $\tau^3$-Retail. These values are then fixed for all test actors.

\subsection{Reference Candidate Mechanisms}
\label{app:candidate_sources}

\noindent \textbf{Environment actions.}
When the environment exposes legal actions, we sample without replacement. In WebShop these are current-page clicks
and search; in ALFWorld they are admissible text commands. 

\noindent \textbf{Offline actions.}
Offline retrieval draws concrete actions observed in branch data under a
compatible decision condition. Retrieved actions are rebound to currently
observed entities and rejected if required identifiers or policy prerequisites
are absent. Retail uses this source to recover useful tool patterns that are
unlikely to appear in a small online sample.

\noindent \textbf{Small-LM actions.}
A small LLM (Qwen2.5-0.5B-Instruct) receives the actor-visible state and proposes alternatives
under several decision modes, such as information gathering, progress,
recovery, and a distinct alternative. The generator never votes on its own actions.

\begin{table}[H]
\centering
\caption{Default candidate source by benchmark and actor family.}
\label{tab:candidate_sources}
\small
\begin{tabular}{lll}
\toprule
Environment & Actor condition & Candidate source \\
\midrule
WebShop & all actors & environment \\
ALFWorld & Qwen3-8B & environment \\
ALFWorld & stronger actors & small LM + environment \\
$\tau^3$-Retail & all actors & small LM + offline \\
\bottomrule
\end{tabular}
\end{table}

\subsection{Baseline Implementations}
\label{app:baseline_implementations}

\noindent\textbf{Self-Reflection.}
The actor reviews each proposed action before execution, including
both user-facing responses and tool calls in Retail.
It performs one conservative self-check and may retain or revise
the proposal. Retaining an action also preserves its original
Thought or response content. Only the final executed action and
its resulting observation enter the persistent interaction history.

\noindent\textbf{Asym-AC.}
We implement Asym-AC with a fixed, zero-shot Qwen3-8B critic.
The critic is invoked at environment-specific intervention points:
purchase actions in WebShop; object manipulation and state-changing actions
in ALFWorld; and database-mutating tool calls in $\tau^3$-Retail.
At each reviewed step, the critic provides one round of free-form feedback,
after which the actor replans once.
The revised action is not reviewed recursively.
These WebShop and ALFWorld intervention points are our benchmark-specific adaptations rather than
settings prescribed by the original work.

\noindent\textbf{AgentRM.}
We implement the AgentRM inference-time search baseline using beam search with
$W_1=W_2=K$ and $K=4$.
At each expansion, retained prefixes are extended with up to $K$ candidate
actions, scored by the released AgentRM reward model, and pruned to the
highest-scoring $K$ prefixes.
The final trajectory is selected according to the AgentRM scores.

\noindent\textbf{Scope of comparison.}
For~\cite{jiang2026asymmetric}, we use the zero-shot Asym-AC variant because no verified author-trained SFT
critic checkpoint was available to us.
We do not directly reproduce the action-conditioned controller of
\citet{zhang2026calibration}, as we could not identify a public implementation
of the controller required to reproduce their intervention setting.

The episode-level budget in Section~\ref{exp:why_COTA_works} is used only as a diagnostic for \name and is not imposed on the baselines in Table~\ref{tab:method_comparison}. Self-Reflection reviews every proposed action before execution. For Asym-AC, we found that per-step criticism performs poorly, so we follow its original design more closely and invoke the critic only at selected consequential decision points.

\subsection{Cost Accounting}
\label{app:cost_details}

\subsubsection{Offline computation}
\label{app:cost_offline_details}

Offline cost includes base rollout generation, branch continuations, and comparator fine-tuning and is reported in Table~\ref{tab:offline_cost}. It is amortized across every actor
and every future evaluation that reuses the comparator, and is therefore kept
separate from the online Avg.~T values in Table~\ref{tab:method_comparison}.

\begin{table}[H]
\centering
\caption{Offline cost accounting. The time unit is H200 GPU-hour.}
\label{tab:offline_cost}
\small
\begin{tabular}{lrrrr}
\toprule
Environment & Base rollout & Branch rollout & Comparator FT & Total GPU-hours \\
\midrule
WebShop          & 1.48 & 78.65 & 1.23 & 81.36 \\
ALFWorld         & 1.97 &  28.39 & 0.80 &  31.15 \\
$\tau^3$-Retail  & 0.27 &   7.79 & 0.36 &   8.42 \\
\bottomrule
\end{tabular}
\end{table}

\subsubsection{Online API-equivalent cost}
\label{app:api_cost_account_rule}
We report API-equivalent inference cost based on the input and output tokens recorded during evaluation. For each model $m$, the cost is computed as
\[
C_m =
\frac{
N^{(m)}_{\mathrm{in}} p^{(m)}_{\mathrm{in}}
+
N^{(m)}_{\mathrm{out}} p^{(m)}_{\mathrm{out}}
}{10^6},
\]
where $N^{(m)}_{\mathrm{in}}$ and $N^{(m)}_{\mathrm{out}}$ denote the total input and output token counts, and $p^{(m)}_{\mathrm{in}}$ and $p^{(m)}_{\mathrm{out}}$ are the corresponding OpenRouter list prices per million tokens. 
We do not apply caching discounts or provider-specific promotions. 
Table~\ref{tab:api_prices} lists the prices used in our accounting.

\begin{table}[H]
\centering
\small
\caption{API prices used for inference-cost accounting. Prices are in USD per million tokens and correspond to the OpenRouter list prices.}
\label{tab:api_prices}
\begin{tabular}{lcc}
\toprule
Model & Input (\$/M) & Output (\$/M) \\
\midrule
Qwen3-8B & 0.117 & 0.455 \\
Qwen3.6-35B-A3B & 0.050 & 0.700 \\
DeepSeek-V4-Flash & 0.068 & 0.168 \\
Llama-3.1-8B-Instruct (AgentRM) & 0.020 & 0.040 \\
\bottomrule
\end{tabular}
\end{table}

AgentRM~\cite{xia2025agentrm} uses fine-tuned Llama-3.1-8B-Instruct as the backbone of its reward model with a scalar prediction head. Since reward-model evaluation scores an input trajectory without autoregressive token generation, we approximate its API-equivalent cost by charging all tokens processed by the reward model at the backbone's input-token rate, with no output-token charge.

\name additionally invokes a 0.5B comparator. We run this comparator locally and do not assign it an external API charge. OpenRouter does not list Qwen2.5-0.5B-Instruct, while a closely matched model at the same scale, Qwen3-0.6B, is available through a free endpoint. We therefore exclude the Qwen2.5-0.5B-Instruct from the reported API-equivalent monetary cost. This convention concerns external API pricing only and does not imply zero underlying hardware cost.

\section{Additional Experimental Results}
\label{app:Moreexperiments}

\subsection{A/B versus A/B/T}
\label{appendix:ABVSABT-exp}

\begin{table}[H]
\centering
\caption{Comparator-target ablation on 100 held-out WebShop tasks. Both rows
use a bidirectional gate and environment-only $K{=}4,R{=}1$ candidates.}
\label{tab:ablation_comparator}
\small
\begin{tabular}{lcccc}
\toprule
Training target & Pref. consistency & Valid output & End reward & Zero reward \\
\midrule
A/B only & 39.51\% & 86.16\% & 0.5031 & 37.0\% \\
A/B/T & \textbf{57.70\%} & \textbf{99.53\%} & \textbf{0.5442} & \textbf{22.0\%} \\
\bottomrule
\end{tabular}
\end{table}

Explicit ties improve order consistency and make abstention available when the
branch outcomes do not justify a winner.

\subsection{Absolute-$Q$ Diagnostics}

\begin{table}[H]
\centering
\caption{Validation quality of the three Qwen2.5-0.5B absolute-$Q$ critics.}
\label{tab:absolute_q_offline}
\small
\begin{tabular}{lccc}
\toprule
Environment & Pairwise ranking acc. (\%) & Spearman & BCE \\
\midrule
WebShop & 69.13 & 0.618 & 0.527 \\
ALFWorld & 64.23 & 0.366 & 0.635 \\
$\tau^3$-Retail & 68.20 & 0.423 & 0.639 \\
\bottomrule
\end{tabular}
\end{table}

\begin{table}[H]
\centering
\caption{Online intervention frequency (\%) for the absolute-$Q$ critic. Forced
reports states where argmax replaces the actor proposal; selective reports gate
rounds that request actor replanning.}
\label{tab:absolute_q_intervention}
\small
\begin{tabular}{lrrrr}
\toprule
& \multicolumn{2}{c}{Qwen3-8B} & \multicolumn{2}{c}{Qwen3.6} \\
\cmidrule(lr){2-3}\cmidrule(lr){4-5}
Environment & Forced & Selective & Forced & Selective \\
\midrule
WebShop & 61.1 & 46.5 & 48.8 & 26.6 \\
ALFWorld & 83.6 & 50.5 & 87.0 & 56.5 \\
$\tau^3$-Retail & 77.7 & 38.7 & 66.0 & 32.9 \\
\bottomrule
\end{tabular}
\end{table}

As shown in~\ref{tab:absolute_q_offline}, the Absolute-$Q$ scalar critic can show reasonable held-out ranking accuracy while remaining
poorly calibrated for online control. 
As shown in~\ref{tab:absolute_q_intervention}, in ALFWorld it frequently scores several
legal but subgoal-irrelevant moves above the proposal $a_t$, causing forced selection
to replace more than 80\% of actor actions. Constructive intervention absorbs some
of these mistakes, but direct pairwise supervision remains substantially more
effective in the closed loop.

\subsection{Cross-Actor Preference Changes}
\label{app:cross_actor}
We test how pairwise action preferences change when the branch-point state and actions are fixed but the continuation actor is replaced. For each environment, we sample 24 held-out prefixes from Qwen3-8B trajectories and replay the same branch-point actions with different continuation actors. This yields 240 sibling action pairs for WebShop, 240 for ALFWorld, and 211 for $\tau^3$-Retail. We first evaluate each pair with Qwen3-8B as the continuation actor and keep only non-tied pairs, leaving 106 pairs on WebShop, 48 on ALFWorld, and 70 on Retail. We then evaluate the same pairs with Qwen3.6-35B-A3B or DeepSeek-V4-Flash. Table~\ref{tab:cross_actor_preference} reports whether the Qwen3 preference is preserved (\emph{Same}), becomes indistinguishable (\emph{Tie}), or changes direction (\emph{Reverse}). Percentages are computed over the non-tied Qwen3 pairs in each environment.

\begin{table}[t]
    \centering
    \small
    \setlength{\tabcolsep}{7pt}
    \begin{tabular}{lcrrr}
        \toprule
        Environment & Actor shift & Same & Tie & Reverse \\
        \midrule
        WebShop
        & Qwen3-8B $\rightarrow$ Qwen3.6-35B-A3B
        & 17.0\%
        & 79.2\%
        & 3.8\% \\

        ALFWorld
        & Qwen3-8B $\rightarrow$ Qwen3.6-35B-A3B
        & 39.6\%
        & 60.4\%
        & 0.0\% \\

        ALFWorld
        & Qwen3-8B $\rightarrow$ DeepSeek-V4-Flash
        & 6.3\%
        & 93.8\%
        & 0.0\% \\

        $\tau^3$-Retail
        & Qwen3-8B $\rightarrow$ Qwen3.6-35B-A3B
        & 42.9\%
        & 55.7\%
        & 1.4\% \\

        $\tau^3$-Retail
        & Qwen3-8B $\rightarrow$ DeepSeek-V4-Flash
        & 44.3\%
        & 52.9\%
        & 2.9\% \\
        \bottomrule
    \end{tabular}
    \caption{
    Changes in pairwise preferences after replacing the Qwen3 continuation actor.
    Each row is computed over action pairs that are non-tied under Qwen3.
    }
    \label{tab:cross_actor_preference}
\end{table}

Most changes are ties rather than reversals. Across the five actor shifts in Table~\ref{tab:cross_actor_preference}, 234 of 342 pair--actor comparisons (68.4\%) become ties, while only 7 (2.0\%) reverse direction. Thus, pairwise preferences are actor-dependent, but replacing the continuation actor more often removes a preference than reverses it.

One interpretation is that a stronger continuation actor can recover from either branch-point action and reach the same outcome, even when the distinction matters for Qwen3-8B. The high tie rates are consistent with this behavior. Accordingly, the cross-actor gains in Table~\ref{tab:method_comparison} should be interpreted as reuse of the \name pipeline across actors, rather than as evidence that the underlying $Q^\pi$ ordering is invariant to the continuation policy. Each branch is evaluated with a single continuation rollout, so this analysis concerns empirical outcome orderings rather than exact $Q^\pi$ values.

\subsection{Effect of Candidate Support: A WebShop Study}
\label{app:candidate_support}

\name can only compare alternatives exposed by the reference candidate mechanism.
We therefore isolate the effect of candidate support on WebShop by varying only this mechanism, while keeping the comparator, intervention rule, candidate budget ($K=4$), and 500 held-out tasks fixed.
Environment and small-LM candidates follow the construction in Section~\ref{sec:implementation} and Appendix~\ref{app:candidate_sources}.
For the additional variants, actor sampling draws high-temperature alternatives from the deployment actor, counterfactual prompting asks the same actor to generate a different next action, and the SFT small-LM uses the same 0.5B backbone fine-tuned on training-set prefix-branch data.

\begin{table}[t]
    \centering
    \small
    \caption{
    Effect of the reference candidate mechanism on WebShop.
    Mean reward and exact success rate are reported on the same 500 held-out tasks.
    }
    \label{tab:candidate_support}
    \begin{tabular}{lcccc}
        \toprule
        & \multicolumn{2}{c}{Qwen3-8B}
        & \multicolumn{2}{c}{Qwen3.6-35B-A3B} \\
        \cmidrule(lr){2-3}\cmidrule(lr){4-5}
        Reference candidate mechanism
        & Mean reward & Success rate
        & Mean reward & Success rate \\
        \midrule

        Environment-only
        & 0.563 & 27.8 \%
        & 0.681 & 56.6 \% \\

        Actor sampling + environment
        & 0.602 & 30.4 \%
        & 0.694 & 57.4 \% \\

        Counterfactual prompting + environment
        & 0.579 & 27.0 \%
        & 0.674 & 56.0 \% \\

        Small-LM sampling + environment
        & 0.609 & 28.8 \%
        & 0.696 & 57.2 \% \\

        SFT small-LM sampling + environment
        & 0.583 & 28.8 \%
        & 0.673 & 54.6 \% \\

        \bottomrule
    \end{tabular}
\end{table}

Table~\ref{tab:candidate_support} shows that the reference candidate mechanism has a measurable effect on \name's performance even when the comparator and intervention rule are fixed.
Generic small-LM generation achieves the highest mean reward for both actors, while direct actor sampling attains the highest exact success rate.
Other mechanisms provide smaller or inconsistent gains: counterfactual prompting offers little improvement over environment-only candidates, and fine-tuning the 0.5B proposer does not further improve performance in this setting.

\section{Prompt Templates}
\label{app:prompts_and_templates_all}

\subsection{Actor Interaction Templates}
\label{app:actor_interaction_templates}

We summarize the actor prompts used in each environment below.
Braced fields denote runtime content, and benchmark-provided policies or
fixed few-shot demonstrations are inserted verbatim where indicated.
We show the common prompt structure for each benchmark and describe
actor-specific deviations separately.

\noindent\textbf{WebShop.}
The Qwen actors receive a fixed shopping instruction with few-shot
demonstrations, followed by the executed history and current page state:

\begin{quote}
\small\ttfamily
[System]\\
You are a shopping agent operating in a text-only WebShop environment.
Follow the user instruction and purchase the correct product.\\
Valid actions are \texttt{search[<keywords>]} and
\texttt{click[<exact text>]}.\\
Reply with exactly:\\
Thought: <brief reasoning>\\
Action: <one executable action>\\
\{fixed few-shot demonstrations\}

[User]\\
Task: \{shopping instruction\}\\
Execution history: \{executed action--observation history\}\\
Current observation: \{current page\}\\
Available actions: \{current executable actions\}
\end{quote}

\noindent
\textbf{Actor-specific variant.}
Qwen3-8B and Qwen3.6-35B-A3B use the rendered execution summary above.
For DeepSeek-V4-Flash, we preserve the archived WebShop prompt and provide
the interaction as the raw ReAct message history instead.

\noindent\textbf{ALFWorld.}
Actors receive the household task, two fixed task-family demonstrations,
and the executed interaction history:

\begin{quote}
\small\ttfamily
[System]\\
You are a household agent operating in the text-only ALFWorld environment.
Complete the task using one currently admissible command per turn.\\
Reply with exactly:\\
Thought: <brief grounded reasoning>\\
Action: <one admissible command>\\
Do not invent entities or issue commands outside the admissible action set.\\
\{two task-family few-shot demonstrations\}

[User]\\
Task: \{household task\}\\
Initial observation: \{initial observation\}\\
\{executed Thought/Action--Observation history\}\\
Current observation: \{current observation\}
\end{quote}

\noindent\textbf{$\tau^3$-Retail.}
Retail follows the benchmark-native customer-service interface.
The system message contains the domain policy, while customer messages,
assistant responses, and tool outputs retain their original API roles:

\begin{quote}
\small\ttfamily
[System]\\
You are a customer-service agent. Follow the supplied policy exactly.
On each turn, either send one natural-language response or make one tool call.
Do not fabricate information, and obtain explicit confirmation before
consequential database operations.\\
<policy>\\
\{benchmark domain policy\}\\
</policy>

[Conversation]\\
\{original user/assistant/tool interaction history\}
\end{quote}

Tool names, descriptions, and argument schemas are provided through the
native API tool interface rather than inserted into the text prompt.
Unlike WebShop and ALFWorld, Retail does not require a textual
\texttt{Thought:}/\texttt{Action:} response format.

\subsection{Comparator Prompt Template}
\label{app:comparator_prompt}

The comparator uses a prompt template across all three environments. Only the rendered state representation is benchmark-specific.
Table~\ref{tab:comparator_state} summarizes the information exposed to the
comparator; in all cases, only executed interaction history is included, and
unexecuted actor reasoning is excluded.

\begin{table}[h]
\centering
\small
\caption{Benchmark-specific state information provided to the comparator.}
\label{tab:comparator_state}
\begin{tabular}{ll}
\toprule
Environment & State representation \\
\midrule
WebShop &
Task, executed history, current page, executable search/click actions \\
ALFWorld &
Task, executed action--observation history, current observation, admissible actions \\
$\tau^3$-Retail &
Policy and interaction state, including grounded entities, confirmations,
tool state, and recent dialogue \\
\bottomrule
\end{tabular}
\end{table}

The resulting state block is inserted into the following shared prompt:

\begin{quote}
\small\ttfamily
[System]\\
You are an expert judge for an interactive agent. Given the task,
observable progress, current state, and two candidate next actions, decide
which action is more likely to lead to correct task completion when the same
base agent continues. If one is clearly better, answer A or B. Otherwise,
answer T.\\

[User]\\
\# Task and observable trajectory\\
\{benchmark-specific state\}\\
\# Candidate actions\\
A: \{action A\}\\
B: \{action B\}\\
\# Answer\\
Reply with exactly one letter: A, B, or T.
\end{quote}

\subsection{Small-LM Candidate Generation}
\label{app:reference_candidate_prompt}

This prompt is used only when the reference candidate mechanism includes
small-LM generation; environment-sampled and offline-retrieved candidates do
not use a generation prompt. The generator produces executable alternatives
but does not score them or decide whether intervention should occur.
We use benchmark-specific generation perspectives to encourage diverse alternatives: product discovery, requirement verification, configuration, and progress in WebShop; subgoal progress, information gathering, recovery, and alternative actions in ALFWorld; and clarification, verification, confirmation, execution, recovery, alternatives, and completion in $\tau^3$-Retail.

\begin{quote}
\small\ttfamily
[Context]\\
\{task, observable interaction history, current state,
and available actions or tools\}\\

[User]\\
Current actor proposal: \{proposed action\}\\
Generation perspective: \{benchmark-specific instruction\}\\

Generate one executable alternative next action from this perspective.
Do not judge whether the current proposal is correct or whether intervention
should occur. Use only information available in the current state, and do not
invent entities, tool arguments, or observations.
Return an action different from the current proposal; if no useful alternative
is available, output NONE.\\

\{benchmark-specific output format\}
\end{quote}

For WebShop and ALFWorld, the generator returns one executable textual action.
For $\tau^3$-Retail, it returns either one user-facing response or one native
tool call with grounded arguments. Exact benchmark-specific formatting is
provided with the corresponding environment implementation.

\subsection{Prompt Templates of Baselines}
\label{app:baseline_prompts}

\subsubsection{Self-Reflection Prompt Template}

The actor receives one conservative self-check before execution. It may keep
its original proposal.

\begin{quote}
\small\ttfamily
[User: Self-check]\\
Unexecuted proposal: \{original Thought and Action, or response/tool call\}\\[4pt]
Review this draft against the task, observed interaction history,
current state, and applicable constraints.
Default to keeping it unless there is a concrete error.
Do not invent a mistake merely to offer a different plan.
An information-gathering action need not already satisfy every
final task requirement.\\[4pt]
Return the action that should actually be executed.
If keeping the original action, preserve its original Thought exactly.
Otherwise, give a grounded reason and one executable action
in the actor's normal format.
\end{quote}

For Retail, keeping a proposal preserves its response text or tool name
and arguments. The prompt explicitly forbids treating a proposed tool
call as completed.

\subsubsection{Asym-AC Critic Prompt Template}

\noindent \textbf{Critic prompt.} The critic reviews the proposed action against observable evidence.
Retail additionally supplies the domain policy and concrete tool
arguments.

\begin{quote}
\small\ttfamily
[System]\\
You are an action-verification critic.
Ground your review in the supplied task, executed interactions,
current observation, available actions, and applicable policy.
Identify concrete errors or confirm that the proposal is appropriate.\\[4pt]
The proposal has NOT been executed.
Missing evidence is not evidence of a violated constraint.
Do not invent commands, requirements, or unobserved state transitions.
Give concise, actionable feedback.\\[4pt]
[User]\\
Task and applicable policy: \{context\}\\
Executed interaction history: \{prefix\}\\
Current observation and available actions: \{state\}\\
Unexecuted actor proposal: \{proposal\}
\end{quote}

\noindent \textbf{Actor revision prompt.}  The actor receives the critic's feedback as:

\begin{quote}
\small\ttfamily
[User: Critic feedback]\\
Unexecuted draft: \{proposal\}\\
Feedback: \{critic feedback\}\\[4pt]
The draft has NOT been executed.
The environment and latest observation are unchanged.
Critic feedback is advice, not an observation or proof of success.
Revise the draft in light of this feedback;
do not skip ahead to the action that would follow it.\\[4pt]
Return the action that should actually be executed in the current state.
You may repeat the original proposal if the feedback confirms it,
preserving its original Thought.
Use the actor's normal output format.
\end{quote}

For Retail, the revision requests one response or native tool call and
explicitly states that no tool result exists for the unexecuted draft.

\subsubsection{AgentRM Scoring Input}

AgentRM receives a serialized interaction prefix rather than a
free-form critique request. Its input layout is summarized below;
the scalar score is read from the reward head, not generated as text.

\begin{quote}
\small\ttfamily
[User]\\
\{environment instructions; policy and tool descriptions for Retail\}\\[4pt]
[Assistant]\\
OK\\[4pt]
[User]\\
\{initial observation or customer request\}\\[4pt]
[Assistant]\\
\{actor response or executed action\}\\[4pt]
[User]\\
\{resulting environment observation or customer response\}\\[4pt]
\ldots\\[4pt]
[Assistant]\\
\{latest response or action in the candidate prefix\}\\[4pt]
[User]\\
\{resulting observation or response\}
\end{quote}

\subsubsection{Absolute-\texorpdfstring{$Q$}{Q} Ablation Prompt Template}

The absolute-$Q$ ablation uses our \name's same observable information as the
pairwise comparator, but scores one state--action pair at a time.

\begin{quote}
\small\ttfamily
[System]\\
You are a process reward model for an interactive agent.
Given the observable task state and a candidate next action, estimate
how likely the action is to lead to successful task completion when
the same actor continues.\\

[User]\\
\# Task and observable state\\
\{benchmark-specific state\}\\
\# Candidate action\\
\{candidate action\}
\end{quote}

The benchmark-specific state renderer follows the same observable-state construction as the comparator. 
The model outputs a scalar score through its classification head, with no
textual rationale.

\subsubsection{\name Feedback Prompt Template}
The comparator never writes free-form criticism. If at least $R$ candidates
win, the actor receives a short, structured message identifying its unexecuted
proposal and listing preferred alternatives in ranked order. The actor is
explicitly allowed to keep the original action or synthesize a third action.

\begin{quote}
\small\ttfamily
[ADVISOR FEEDBACK]\\
The proposed action \{proposal\} may not be the best next step.
Preferred alternatives: 1. \{winner 1\}; 2. \{winner 2\}.\\
Reconsider the current state and output the action that should actually be executed. The recommendations are non-binding.
\end{quote}

\end{document}